\documentclass[preprint,12pt]{elsarticle}

\usepackage{amssymb}
\usepackage{url}
\usepackage{amsmath}
\usepackage{amssymb,amsfonts}
\usepackage{graphicx}
\usepackage{textcomp}
\usepackage{xcolor}
\usepackage{algorithm} 
\usepackage{algpseudocode}
\usepackage{multirow}
\usepackage{booktabs}
\usepackage{acronym}

\journal{Knowledge-based Systems}

\begin{document}

\begin{frontmatter}



\title{Mixup Domain Adaptations for Dynamic Remaining Useful Life Predictions}


\author[inst1]{Muhammad Furqon\fnref{fn1}}

\affiliation[inst1]{organization={STEM, University of South Australia},
            addressline={Mawson Lakes Campus}, 
            state={SA},
            country={Australia}}

\author[inst1]{Mahardhika Pratama\fnref{fn1}}
\author[inst1]{Lin Liu}
\author[inst1]{Habibullah Habibullah}
\author[inst1]{Kutluyil Dogancay}

\fntext[fn1]{M. Furqon and M. Pratama share equal contributions.}

\begin{abstract}
Remaining Useful Life (RUL) predictions play vital role for asset planning and maintenance leading to many benefits to industries such as reduced downtime, low maintenance costs, etc. Although various efforts have been devoted to study this topic, most existing works are restricted for i.i.d conditions assuming the same condition of the training phase and the deployment phase. This paper proposes a solution to this problem where a mix-up domain adaptation (MDAN) is put forward. MDAN encompasses a three-staged mechanism where the mix-up strategy is not only performed to regularize the source and target domains but also applied to establish an intermediate mix-up domain where the source and target domains are aligned. The self-supervised learning strategy is implemented to prevent the supervision collapse problem. Rigorous evaluations have been performed where MDAN is compared to recently published works for dynamic RUL predictions. MDAN outperforms its counterparts with substantial margins in 12 out of 12 cases. In addition, MDAN is evaluated with the bearing machine dataset where it beats prior art with significant gaps in 8 of 12 cases. Source codes of MDAN are made publicly available in \url{https://github.com/furqon3009/MDAN}.
\end{abstract}


\begin{highlights}
\item \textcolor{black}{We propose mix-up domain adaptation for time-series unsupervised domain adaptation.}
\item \textcolor{black}{MDAN is applied to dynamic remaining useful life predictions and fault diagnosis.}
\item \textcolor{black}{We propose a self-supervised learning method via a controlled reconstruction learning.} 
\end{highlights}


\end{frontmatter}


\section{Introduction}
\label{sec:intro}
Predictions of remaining useful life (RUL) plays vital role for predictive maintenance. Reliable RUL predictions enable maximum utilization of assets leading to improved productivity and low maintenance costs with minimum risks of catastrophic failures \cite{Vachtsevanos2006IntelligentFD}. There exist two main approaches for RUL estimations \cite{Ragab2021ContrastiveAD,Ragab2020AdversarialTL}: model-based approaches and data-driven approaches. Model-based approaches rely on physic-based approaches to model non-linear dynamics of a system and to link it to the fault progression \cite{Miao2019JointLO}. Although such analytical approach is highly accurate, such approach is cumbersome to be applied to highly complex systems and problem-specific so that it cannot be easily modified to different but related problems. Data-driven approaches \cite{Jiang2019RecentAI} as an alternative utilize available data and artificial intelligence (AI) techniques to craft black-box models unveiling hidden relationships between input and output variables. Compared to the first approach, the second approach is fast to adopt and significantly less expensive than the first one because data samples can be conveniently collected from different sensors. 

The advent of AI techniques have driven recent progresses of RUL predictions where consistent accuracy improvements are observed. The early attempt involves a two-staged approach where at first input features are extracted across different sensor readings followed by the applications of classic machine learning variants such as support vector machines, artificial neural networks, etc \cite{Khan2018ARO}. The feature extraction step usually involves in-depth domain knowledge and problem-specific thus hindering its applicability to a wide range of processes \cite{Ragab2020AdversarialTL,Ragab2021ContrastiveAD}. This issue leads to the use of deep learning techniques which support natural feature engineering mechanisms through the end-to-end training processes. Several deep learning approaches \cite{Li2018RemainingUL,Zhu2019EstimationOB,Deutsch2018UsingDL,Ma2018DeepCA,Huang2019ABL,Chen2021MachineRU} have been used for RUL estimations and offer performance gains compared to the two-staged approaches.  

Despite recent progresses of data-driven approaches for RUL predictions, existing approaches rely on the I.I.D condition which hinder their deployments in dynamic conditions because predictions of old models are outdated. \textcolor{black}{That is, training and testing data are in practice drawn from the different distributions, i.e., machine or asset environments are changing due to different parameters or aging conditions.} Naive solution to this problem is to retrain the predictive model from scratch using new and old data. Such approaches usually call for fully labelled samples obtained via the run-to-failure tests \cite{Ragab2020AdversarialTL,Ragab2021ContrastiveAD} imposing considerable labelling costs. \cite{Ragab2020AdversarialTL,Ragab2021ContrastiveAD} resolve this problem via algorithmic development of unsupervised domain adaptation methods for RUL predictions handling fully labelled source domain and unlabelled target domain. Their approaches are based on popular adversarial domain adaptation techniques originally proposed for image classification problems \cite{Ganin2015DomainAdversarialTO}. Such approaches require a discriminator network aside from the main network and thus impose prohibitive space and memory complexities. Their approaches also ignore discriminative information of the target domain. In addition, this paper distinguishes itself from these works where the mix-up domain adaptation technique is developed.

\begin{figure}
    \centering
    \includegraphics[scale=1]{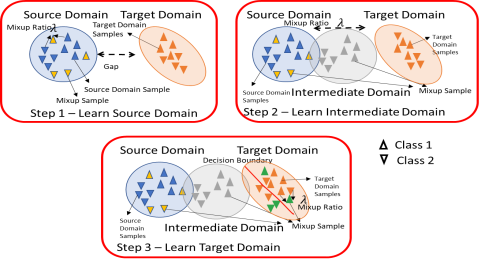}
    \caption{\textcolor{black}{Visualization of Mixup Domain Adaptation (MDAN): The first step learns the source domain using original source samples and mixup samples. An intermediate domain is established in the second step using the mix-up samples which linearly interpolates source and target samples. The last step is to extract the discriminative information of the target domain using pseudo labels and the predictive consistencies using mixup samples. Note that the mixup ratio controls the distributions of the mixed samples}}
    \label{fig:my_label}
\end{figure}

A mix-up domain adaptation (MDAN) method is proposed in this paper for dynamic RUL predictions where, given a labelled source domain without any labels for a target domain with different distributions of the two domains,  a predictive model is created to generalize well to the target domain. MDAN is based on a three-staged learning strategy where a model is at first trained to master the source domain under the mix-up regularization strategy \cite{Wu2020DualMR} to enhance predictive consistency and to enrich intrinsic structures of the latent space. The second stage is designed to create an intermediate mix-up domain where the source and target domains are aligned onto it. This is made possible by applying the progressive mix-up approach \cite{zhu2023progressive} which creates convex combinations between the source-domain samples and the target-domain samples using pseudo-labels of the target domain. That is, the key lies in the mix-up ratio assigned as the Wassertein distance of the source-to-mixup domain and the target-to-mixup domain. The last step is to apply the mix-up regularization strategy in the target domain using pseudo-label information. The goal is to extract discriminative information of the target domain where label information do not exist. Using pseudo labels directly is noisy and thus the mix-up strategy is performed in the target domain as a consistency regularization approach. The self-supervised learning strategy via the controlled reconstruction learning step \cite{Chowdhury2022TARNetTR} is integrated to prevent the supervision collapse problems and to create transferrable representations. The overview of MDAN is pictorially illustrated in Fig. \ref{fig:my_label}.

This paper conveys at least five major contributions: 
\begin{itemize}
    \item the mix-up domain adaptation (MDAN) is proposed where it consists of three mix-up mechanisms in the source, intermediate and target domains. The first step is to learn discriminative information of labelled source domain while the second step is to align both source and target domains onto the intermediate domain. The last step is to drive adaptable representations of the intermediate domain back to the target domain. To the best of our knowledge, we are the first to put forward the mixup domain adaptation strategy for time-series problems;
    \item this paper offers an extension of the progressive mix-up strategy \cite{zhu2023progressive} and the mix-up regularization strategy \cite{Wu2020DualMR}. {\cite{Wu2020DualMR} is designed for image classification problems while our approach here targets the RUL prediction problems using time-series information of sensors. \cite{zhu2023progressive} still assumes the presence of few labelled samples of the target domain and works for image classification problems whereas ours do not assume any labels of the target domain and focuses on the regression problems rather than the classification problems. Note that the time-series problem is fundamentally different from the image classification problem because of the presence of temporal dependency of time-series data. In addition, \cite{Wu2020DualMR} still relies on the adversarial learning strategy which does not exist in MDAN whereas \cite{zhu2023progressive} only concerns creations of intermediate domains via the Mixup technique,i.e., there does not exist any mixup strategies in the source and target domains which can boost the performance};
    \item a self-supervised learning strategy via the controlled reconstruction learning step \cite{Chowdhury2022TARNetTR} is devised to improve transferability of source and target features. {Note that \cite{Chowdhury2022TARNetTR} does not yet explore the controlled reconstruction technique for domain adaptations. Such approach is used for the pre-training phase which does not address any dynamic situations. On the other hand, we combine such technique for learning source domain samples to improve their transferability};
    \item the advantage of MDAN has been rigorously evaluated for RUL predictions of real-world turbofan engine datasets. It is also compared with prominent algorithms where MDAN outperforms those algorithms with significant margins in 12 out of 12 domain adaptation cases. By extension, we also test the performance of MDAN with the bearing machine dataset in which MDAN is superior to its competitor in 8 of 12 cases;
    \item the source codes of MDAN are made publicly available in \url{https://github.com/furqon3009/MDAN} to assure reproducible results and convenient further studies.
\end{itemize} 

\section{Related Works}
\subsection{Mixup}
Mix-up \cite{Zhang2017mixupBE} constitutes a data augmentation technique which linearly interpolates two points, thus resulting in their convex combinations. It has wide applications encompassing semi-supervised learning, domain adaptation, domain generalization, etc. In \cite{Wu2020DualMR}, the mixup method has been applied for unsupervised domain adaptation (UDA) combined with the adversarial domain adaptation technique. That is, the class-level and domain-level mix-up approaches are proposed as a consistency regularization approach as well as enrichment of latent space representations. \cite{Shu2021OpenDG} implements the mix-up strategy for domain generalization where the source domain is augmented. The two approaches depend on a pre-defined mix-up ratio, e.g., beta distribution, which is not sufficiently flexible to dampen the domain discrepancies. As a result, the mix-up approach is combined with other domain adaptation techniques to create domain-invariant representations. In \cite{Mai2019MetaMixUpLA}, the mix-up ratio is dynamically adjusted via the meta-learning technique. This approach confirms the importance of mix-up ratio but excludes yet the case of domain alignment. \cite{zhu2023progressive} presents the cross-domain adaptation with the mix-up method where such approach is used to establish the intermediate domain of the source and target domain by progressively formulating the mix-up ratio as the Wasserstein distance. This approach assumes the presence of labelled samples of the target domain and is devised for image problems. Our approach extends this approach for UDA cases using time-series data. \textcolor{black}{\cite{Zhu2023PatchMixTF} proposes a patchmix method seen as an extension of mix-up for UDA problems with the transformer model. The concept of mix-up is applied for multi-target domain adaptation problems using the 3D data \cite{Sinha2023MEnsAME}. \cite{Li2023AUU} uses a combination between the mix-up regularization technique and the adversarial learning strategy for UDA problems of myoelectric data.} 

\subsection{Deep Learning for Predictive Maintenance}
The advent of deep learning technologies have attracted research attention in the predictive maintenance community whose automatic feature engineering property via the end-to-end learning process sidesteps laborious feature extraction processes \cite{Khan2018ARO}. \cite{Zhu2019EstimationOB} applies the CNN for RUL predictions.  A joint loss is proposed in \cite{Liu2020SimultaneousBF} for fault detection and RUL estimations. \cite{Deutsch2018UsingDL} puts forward DBN for RUL predictions. The bi-directional LSTM is adopted in \cite{Huang2019ABL} to predict RUL under different operating conditions. \cite{Chen2021MachineRU} develops an attention-based LSTM approach for RUL predictions. These approaches do not yet consider the case of dynamic RUL predictions where there exist shifts in data distributions between the training mode and the deployment mode.

\subsection{Domain Adaptations for RUL Predictions}
Unsupervised domain adaptation (UDA) plays vital role for RUL predictions to resolve dynamic situations due to changing operational conditions, aging equipment and its parts, etc. UDA paves a way to adapt a model to new conditions efficiently with the absence of any labelled samples. In \cite{Costa2019RemainingUL}, a LSTM network is combined with the domain adaptation neural network (DANN) \cite{Ganin2015DomainAdversarialTO} to achieve the domain-invariant condition. \cite{Li2019DomainAR} performs the UDA mechanism using the batch normalization (BN) technique. \cite{Ragab2020AdversarialTL} proposes adversarial domain adaptations for RUL predictions and is extended in \cite{Ragab2021ContrastiveAD} integrating the self-supervised learning approach. \cite{Li2022DomainAR} makes use of the transformer as a base model and adds the semantic discriminator for domain adaptations while \cite{Cheng2021DeepTL} proposes dynamic domain adaptation network and dynamic adversarial adaptation network for RUL estimations of mechanical bearings under different working conditions. This paper considers the same problem here as per \cite{Li2019DomainAR,Li2022DomainAR,Ragab2020AdversarialTL,Ragab2021ContrastiveAD,Cheng2021DeepTL} with different approaches where a non-adversarial training approach is developed. It is worth-mentioning that the adversarial-based approach is steered by a domain discriminator imposing extra space and memory complexities. \textcolor{black}{The concept of UDA has also experienced significant progress in the EEG data domain \cite{Zhang2022MultiModalityF,Jiang2021EEGBasedDD}. In \cite{Jiang2021EEGBasedDD}, the UDA method based on the TSK fuzzy system is proposed while \cite{Zhang2017mixupBE} proposes a multi-kernel learning for UDA with the EEG data.}

\section{Preliminaries}
\subsection{Problem Formulation}
Unsupervised Domain Adaptation (UDA) is considered here where a model is provided with a fully labelled source domain and a unlabelled target domain. That is, a source domain carries tuples of data points $\mathcal{S}=\{(x_{i}^{S},y_{i}^{S})\}_{i=1}^{n_s}$ where $n_s$ is the number of labelled samples, $x^{S}\in\mathcal{X}_{S}$, $y^{S}\in\mathcal{Y}_{S}$ and $\mathcal{X}_{S}\times\mathcal{Y}_{S}\backsim\mathcal{D}_{S}$. A target domain suffers from the absence of labelled samples $\mathcal{T}=\{(x_i^{T})\}_{i=1}^{n_t}$ and the covariate shift problem whose marginal distribution differs from that of the source domain $P_{S}(X)\neq P_{T}(X)$, thus following different distributions. $n_t$ stands for the number of samples in the target domain, $x^{T}\in\mathcal{X}_{T},y^{T}\in\mathcal{Y}_{T}$ and $\mathcal{X}_{T}\times\mathcal{Y}_{T}\backsim\mathcal{D}_{T}$. The cross domain problem $\mathcal{D}_{T}\neq\mathcal{D}_{S}$ is resulted from different marginal distributions. The source domain and the target domain share the same input space and the label space $\mathcal{X}_{S}=\mathcal{X}_{T}$, $\mathcal{Y}_{S}=\mathcal{Y}_{T}$. The RUL prediction problem constitutes the multivariate time-series problem $x_{i}^{S},x_{i}^{T}\in\Re^{M\times K}$ where $M,K$ respectively stand for the number of sensors and the number of time steps. 

The goal of this problem is to craft a predictive model $f_{\phi}(g_{\psi}(.))$ to perform well to the unlabelled target domain $\mathcal{D}_{T}$ by transferring labelled information of the source domain $\mathcal{D}_{S}$ where $g_{\psi}:\mathcal{X}\rightarrow \mathcal{G}$ is a feature extractor parameterized by $\psi$ and $f_{\phi}:\mathcal{G}\rightarrow\mathcal{Y}$ is a predictor network parameterized by $\phi$. This is attained by learning the source domain and by minimizing the discrepancies of the source and target domains $\mathcal{L}\triangleq\mathbb{E}_{(x,y)\backsim\mathcal{D}_{S}}[l(f_{\phi}(g_{\psi}(x)),y)]+d(\mathcal{D}_{S},\mathcal{D}_{T})$ where $l(.)$ is usually formulated as the mean-squared error (MSE) loss function. 

\subsection{Mix-up}
Mix-up can be considered as regularization approach and shown as a lower bound of the Lipschitz constant of the gradient of the neural network \cite{Mai2019MetaMixUpLA}. Suppose $f:\Re^{d}\rightarrow\Re$ is a differentiable function and its gradient is $\kappa$-Lipschitz continuous:
\begin{equation}\label{mixup1}
    \forall x, x^{'}\Re^{d}\quad||\nabla f(x)-\nabla f(x^{'})||\leq||x-x^{'}||
\end{equation}
the following inequality is considered:
\begin{equation}\label{mixup2}
    \begin{split}
        |f(\lambda x+(1-\lambda x^{'}))-[\lambda f(x)+(1-\lambda)f(x^{'})]|\\\leq\frac{\lambda(1-\lambda)\kappa}{2}||x-x^{'}||^{2}
    \end{split}
\end{equation}
where $\lambda\in[0,1]$. We can see $x$ and $x^{'}$ in \eqref{mixup2} as $x_i$ and $x_j$ in the mixup scheme. The relationship between the mixup technique and the gradient Lipschitz continuity is derived in the following proposition. 

\textit{Proposition 1 (Link Between mixup and gradient Lipschitz continuity) \cite{Mai2019MetaMixUpLA}:} for all $x$ and $x^{'}$ in $\Re^{d}$, the following is derived.
\begin{equation}
    \begin{split}
        f(\lambda x + (1-\lambda)x^{'})\\
        =f(x^{'})+\lambda\int_{0}^{1}(\nabla f(\lambda t x + (1-\lambda t)x^{'}),x-x^{'})dt\\
        =f(x^{'})+\lambda[f(x) - f(x^{'})]\\
        +\lambda[\int_{0}^{1}(\nabla f(\lambda tx+(1-\lambda t)),x-x^{'})dt-(f(x)-f(x^{'}))]
    \end{split}
\end{equation}
The following condition holds. 
\begin{equation}
\begin{split}
        |f(\lambda x+(1-\lambda)x^{'})
        -(\lambda f(x)+(1-\lambda)f(x^{'}))|
        =\lambda|\int_{0}^{1}(\nabla f(\lambda tx+(1-\lambda t)x^{'})\\
         -\nabla f(tx+(1-t)x^{'}),x-x^{'})dt|\\
         \leq\lambda\int_{0}^{1}|(\nabla f(\lambda tx+(1-\lambda t)x^{'})\\
         -\nabla f(tx+(1-t)x^{'}),x-x^{'})|dt\\
         \leq\lambda\int_{0}^{1}||\nabla f(\lambda tx+(1-\lambda t)x^{'})\\
         -\nabla f(tx+(1-t)x^{'})||||x-x^{'}||\\ \leq\lambda\int_{0}^{1}(1-\lambda)t\kappa||x-x^{'}||^{2}dt\\
        =\frac{\lambda(1-\lambda)\kappa}{2}||x-x^{'}||^{2}
\end{split}
\end{equation}
where the second inequality is derived from the Cauchy-Schwartz inequality and the third inequality originates from \eqref{mixup1}. Overall, this equation suggests the importance of the mixup ratio $\lambda$ in order to control the Lipschitz constant. This fact is obvious when $x$ is distant from $x^{'}$.  

\begin{algorithm}
	\caption{MDAN}
 \hspace*{\algorithmicindent} \textbf{Input:}  Source domain: \textit{\(S=\left \{x_{i}^{S},y_{i}^{S} \right \}_{i=1}^{n_{s}}\)  } \\
  \hspace*{\algorithmicindent} Target domain: \(T=\left \{ x_{i}^{T} \right \}_{i=1}^{n_{t}}\) \\
 \hspace*{\algorithmicindent} \textbf{Output:} Configuration of \textbf{MDAN} \\
	\begin{algorithmic}[1]
		\For {number of iterations}
				\State $\left (  x^{S},y^{S} \right )\leftarrow RANDOMSAMPLE\left ( X_{S},Y_{S} \right ) $
    			\State $\left ( x^{T} \right )\leftarrow RANDOMSAMPLE\left ( X_{T} \right )$
                \State $\lambda \leftarrow RANDOMSAMPLE\left ( Beta\left ( \alpha ,\alpha  \right ) \right )$
			\State $\#Source Domain Training$
			\State Compute source loss $\mathcal{L}^{or}_{S} \leftarrow eq.\left ( 5 \right )$ 
			\State Get source feature-level mixup and label mixup $\left ( \widetilde{g}_{\psi }\left ( x \right ),\widetilde{y} \right )\leftarrow eq.\left ( 6 \right )$
            \State Compute source mixup loss $\mathcal{L}_{S}^{mx}\leftarrow eq. \left ( 7 \right )$
            \State $m_{t}\leftarrow RANDOMSAMPLE\left ( 0,1 \right )$
            \State Compute source mask, unmask, and reconstruction loss $\left ( \mathcal{L}_{m},\mathcal{L}_{um},\mathcal{L}_{R} \right )\leftarrow eq. \left ( 8,9,10 \right )$
            \State Compute source domain loss $\mathcal{L}_{S}\leftarrow eq. \left ( 11 \right )$
            \State $\#Intermediate Domain Training$
            \State Get pseudo label $\left ( \widehat{y}_{i}\right ) \leftarrow f_{\phi }\left ( g_{\psi }\left ( x_{i} \right ) \right ) $
            \State Get intermediate data-label mixup $\left ( \widetilde{x},\widetilde{y} \right )$ and feature-level mixup $\left ( \widetilde{g}_{\psi }\left ( x \right )\right )\leftarrow eq.\left ( 12,13 \right )$
            \State Compute intermediate domain loss $\mathcal{L}_{cd}\leftarrow eq. \left ( 17 \right )$
            \State $\#Target Domain Training$
            \State Compute target loss $\mathcal{L}^{or}_{T} \leftarrow eq.\left ( 19 \right )$
            \State Compute target mixup loss $\mathcal{L}^{mx}_{T} \leftarrow eq.\left ( 20 \right )$
            \State Compute target domain loss $\mathcal{L}_{T} \leftarrow eq.\left ( 21 \right )$
		\EndFor
	\end{algorithmic} 
\end{algorithm}

\section{Learning Policy of MDAN}
MDAN learning policy comprises three steps: source domain training, intermediate domain training and target domain training. The source domain training process learns original labelled samples and mixup samples in the supervised learning manner while also performing the self-supervised training process with the absence of any labels via the controlled reconstruction process. The intermediate domain training process is executed by creating mixup samples between the source domain samples and the target domain samples aided by the pseudo-labelling steps in both input level and feature level. The mixup samples are learned in the supervised learning manner. Last but not least, the target domain training process is driven by the self-learning mechanism of unlabelled target domain samples and the mixup regularization strategy to combat the issue of noisy pseudo labels.
\subsection{Source Domain}
\begin{figure}
    \centering
    \includegraphics[width=1\linewidth]{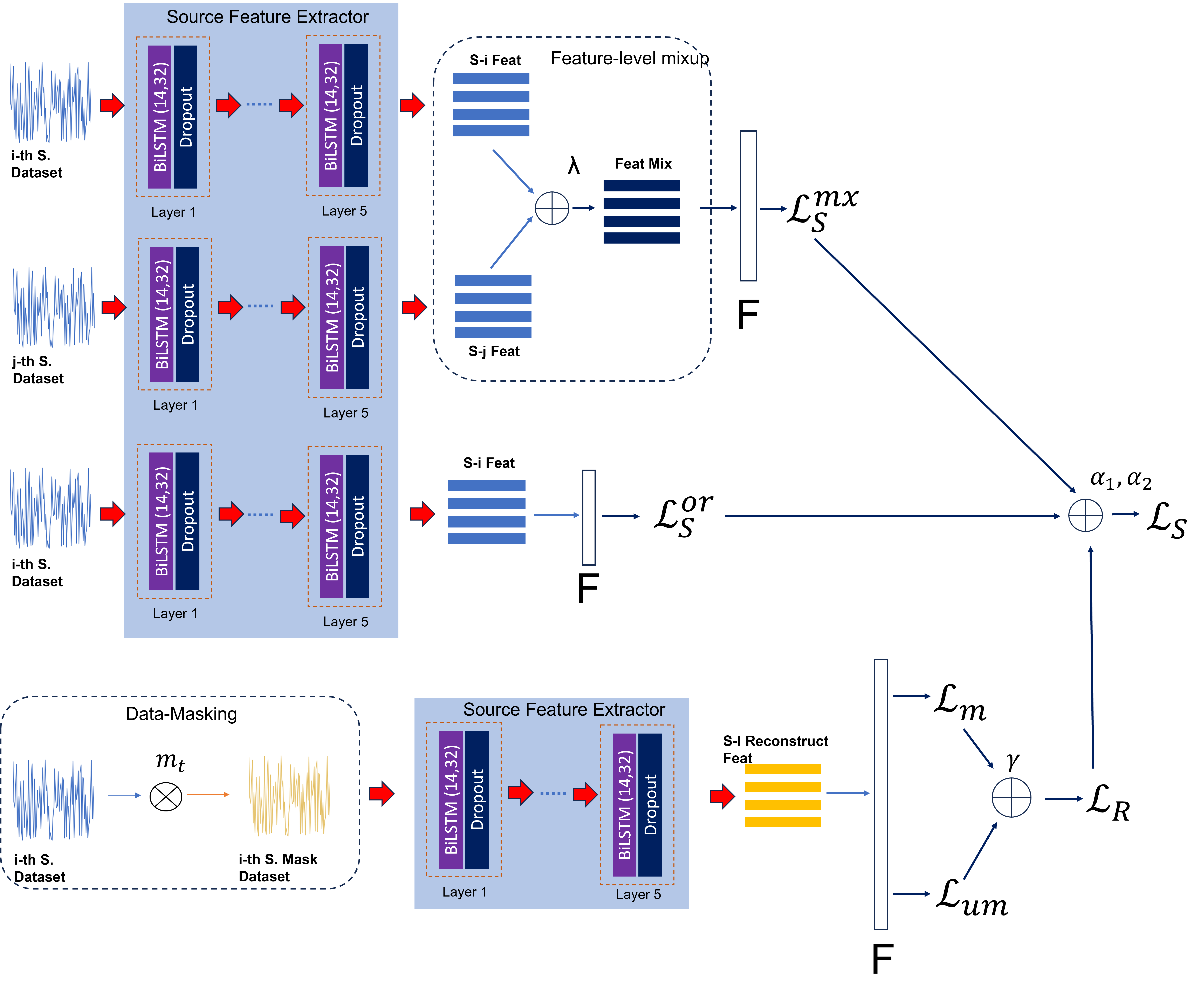}
    \caption{\textcolor{black}{Source Domain Training: a model is trained to minimize the supervised learning loss, the mix-up loss and the self-supervised loss. The supervised learning loss is formulated as a MSE loss function while the mix-up mechanism is done in the feature level. The mixed samples and their corresponding labels are learned using the MSE loss function. The self-supervised loss is formulated as a controlled reconstruction learning process where the main goal is to reconstruct the masked inputs. }}
    \label{fig:source domain}
\end{figure}

\begin{figure}
    \centering
    \includegraphics[width=0.8\linewidth]{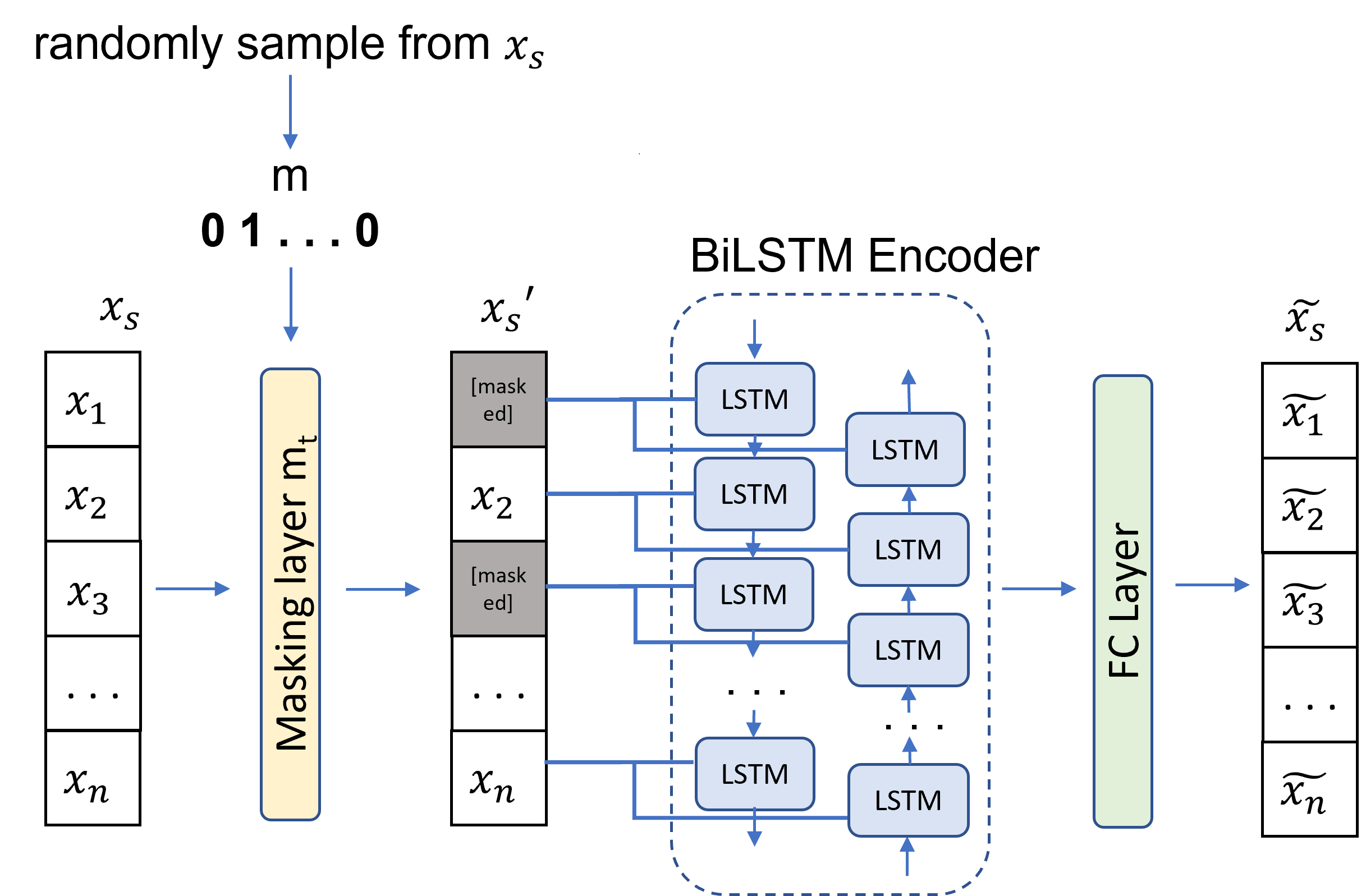}
    \caption{\textcolor{black}{Self-Supervised Learning: This phase is formulated as a controlled reconstruction learning process of randomly masked input samples. The goal is to predict the original input attributes given masked input features. The reconstruction process is done via the backbone network with the biLSTM encoder and the fully connected layer predictor.}}
    \label{fig:reconstruct}
\end{figure}

MDAN's learning procedure starts with the labelled source domain in which samples are learned in the supervised manner using the MSE loss functions:
\begin{equation}
\mathcal{L}_{S}^{or}=\mathbb{E}_{(x,y)\backsim\mathcal{D}_{S}}[l(f_{\phi}(g_{\psi}(x)),y)]
\end{equation}
where $l(.)$ is defined as the MSE loss function. The mixup strategy is implemented to construct augmented samples which happen to be convex combinations of a pair of samples $(x_i,y_i)$ and $(x_j,y_j)$. Unlike the original version taking place in the input space, the mixup technique is undertaken in the manifold level to enrich latent representations. 
\begin{equation}
    \begin{split}
        \tilde{g}_{\psi}(x)=\lambda g_{\psi}(x_i^{S})+(1-\lambda)g_{\psi}(x_j^{S})\\
        \tilde{y}=\lambda y_i^{S} + (1-\lambda)y_j^{S}
    \end{split}
\end{equation}
where $\lambda$ is the mixup coefficient selected randomly from the beta distribution $Beta(\alpha,\alpha),\alpha\in(0,\infty)$. The mixup samples are learned in the same manner as per the original samples:
\begin{equation}
 \mathcal{L}_{S}^{mx}=\mathbb{E}_{(x,y)\backsim\mathcal{D}_{S}}[l(f_{\phi}(\tilde{g}_{\psi}(x)),\tilde{y})]
\end{equation}
This strategy is seen as the consistency regularization mechanism since the mixup approach performs linear interpolations. In addition, the self-supervised learning strategy is incorporated to induce transferable features. This is done via the controlled reconstruction learning process \cite{Chowdhury2022TARNetTR}. Given $m_t$ as a binary mask at the $t-th$ time step, a time-series input samples $x_t$ is zeroed if $m_t=1$. The controlled reconstruction loss function comprises the masked loss function and the unmasked loss function as follows:
\begin{equation}
    \mathcal{L}_{m}=\mathbb{E}_{(x)\backsim\mathcal{D}_{S}}[m_tl(f_{\phi}(g_{\psi}(x)),x)]
\end{equation}
\begin{equation}
    \mathcal{L}_{um}=\mathbb{E}_{(x)\backsim\mathcal{D}_{S}}[(1-m_t)l(f_{\phi}(g_{\psi}(x)),x)]
\end{equation}
Since the reconstruction process of the masked input depends on the reconstruction process of the unmasked input, both masked and unmasked loss functions are combined with a tradeoff parameter $\gamma\in[0,1]$:
\begin{equation}
    \mathcal{L}_{R}=\gamma\mathcal{L}_{m}+(1-\gamma)\mathcal{L}_{um}
\end{equation}
where $m_t$ is randomly sampled. The learning process of the source domain is defined as follows:
\begin{equation}
    \min_{\phi,\psi} \mathcal{L}_{S}^{or}+\alpha_1\mathcal{L}_{S}^{mx}+\alpha_2\mathcal{L}_{R}
\end{equation}
where $\alpha_1,\alpha_2$ are tradeoff constants. Fig. \ref{fig:source domain} visualizes the learning strategies of the source domain and Fig. \ref{fig:reconstruct} portrays the self-supervised learning strategy. 

\subsection{Intermediate Mixup Domain}
An intermediate mixup domain is established using the mixup method by linearly interpolating the source-domain sample and the target-domain sample \cite{zhu2023progressive}. This procedure initiates with the pseudo-labelling step to produce $\hat{y}_{i}=f_{\phi}(g_{\psi}(x_i))$ for target-domain samples whose target samples are not available. The pseudo-labelling step might induce noisy pseudo labels undermining the generalization performance. The mix-up regularization strategy is applied to minimize the adverse effects of noisy pseudo labels. The mix-up mechanism is carried out for $(x_i^{S},y_{i}^{S})$ and $(x_i^{T},\hat{y}_i^{T})$ in both input and feature levels to enrich both data and feature diversities.  
\begin{equation}
    \begin{split}
        \tilde{x}=\lambda x_{i}^{S}+(1-\lambda)x_{i}^{T}\\
        \tilde{y}=\lambda y_{i}^{S}+(1-\lambda)\hat{y}_{i}^{T}
    \end{split}
\end{equation}
where $\lambda$ is the mix-up ratio. Feature-level mix-up between $(g_{\psi}(x_i^{S}),y_i)$ and $(g_{\psi}(x_i^{T}),\hat{y}_i)$ allows intermediate features and direct interactions with the predictor $f_{\phi}(.)$.
\begin{equation}
    \begin{split}
        \tilde{g}_{\psi}(x)=\lambda g_{\psi}(x_i^{S})+(1-\lambda)g_{\psi}(x_i^{T})\\
        \tilde{y}=\lambda y_{i}^{S}+(1-\lambda)\hat{y}_{i}^{T}
    \end{split}
\end{equation}
where $\lambda$ is the mix-up ratio and akin to that of the input level. Note that a large $\lambda$ leads to closer-to-source samples whereas a small $\lambda$ results in closer-to-target samples. Since the sample-level mix-up samples occupy the same feature space as the feature-level mixup samples, their classification decisions should be the same. Their joint use is equivalent to the penalty.

The mixup ratio plays vital role to control the interpolation intensity which determines the knowledge transfer \cite{zhu2023progressive}. We follow the progressive sampling strategy \cite{zhu2023progressive} where the Wasserstein distance is applied to measure the discrepancies of the source to intermediate domain $d(\mathcal{D}_{S}, \mathcal{D}_{mix})$ and the target to intermediate domain $d(\mathcal{D}_{T}, \mathcal{D}_{mix})$. That is, initially, the intermediate domain is introduced as closer to the target domain than the source domain. The mix-up ratio is gradually adjusted to push the mixup domain to be closer to the source domain, thereby achieving domain alignments. This strategy implies reductions of large domain gaps by initiating close to the target and moving toward to the source, ensuring knowledge transfer seamlessly. This is achieved by introducing a weighting factor $q$ to depict the similarity of the source domain.
\begin{equation}
    q=\exp{(-\frac{d(\mathcal{D}_{S},\mathcal{D}_{mix})}{(d(\mathcal{D}_{S},\mathcal{D}_{mix})+d(\mathcal{D}_{T},\mathcal{D}_{mix})T)})}
\end{equation}
where $T$ is the temperature constant set as 0.05. $q$ is initially expected to be small and applied to $\lambda$ in the moving average manner. 
\begin{equation}
    \lambda_{n}=\frac{n(1-q)}{N}+q\lambda_{n-1}
\end{equation}
where $N$ is the total number of iterations and $n$ is the current iteration index. The uniform distribution $U$ along with the random perturbation is introduced to stabilize the training process. 
\begin{equation}
    \tilde{\lambda}_{n}=Clamp(U(\lambda_{n}-\sigma,\lambda_{n}+\sigma),min=0.0,max=1.0)
\end{equation}
where $\sigma$ is a local perturbation range fixed at 0.2. $\tilde{\lambda}_{n}$ is sampled from uniform distribution and clamped in $\tilde{\lambda}_{n}\in[0,1]$. 

\begin{figure}
    \centering
    \includegraphics[width=1\linewidth]{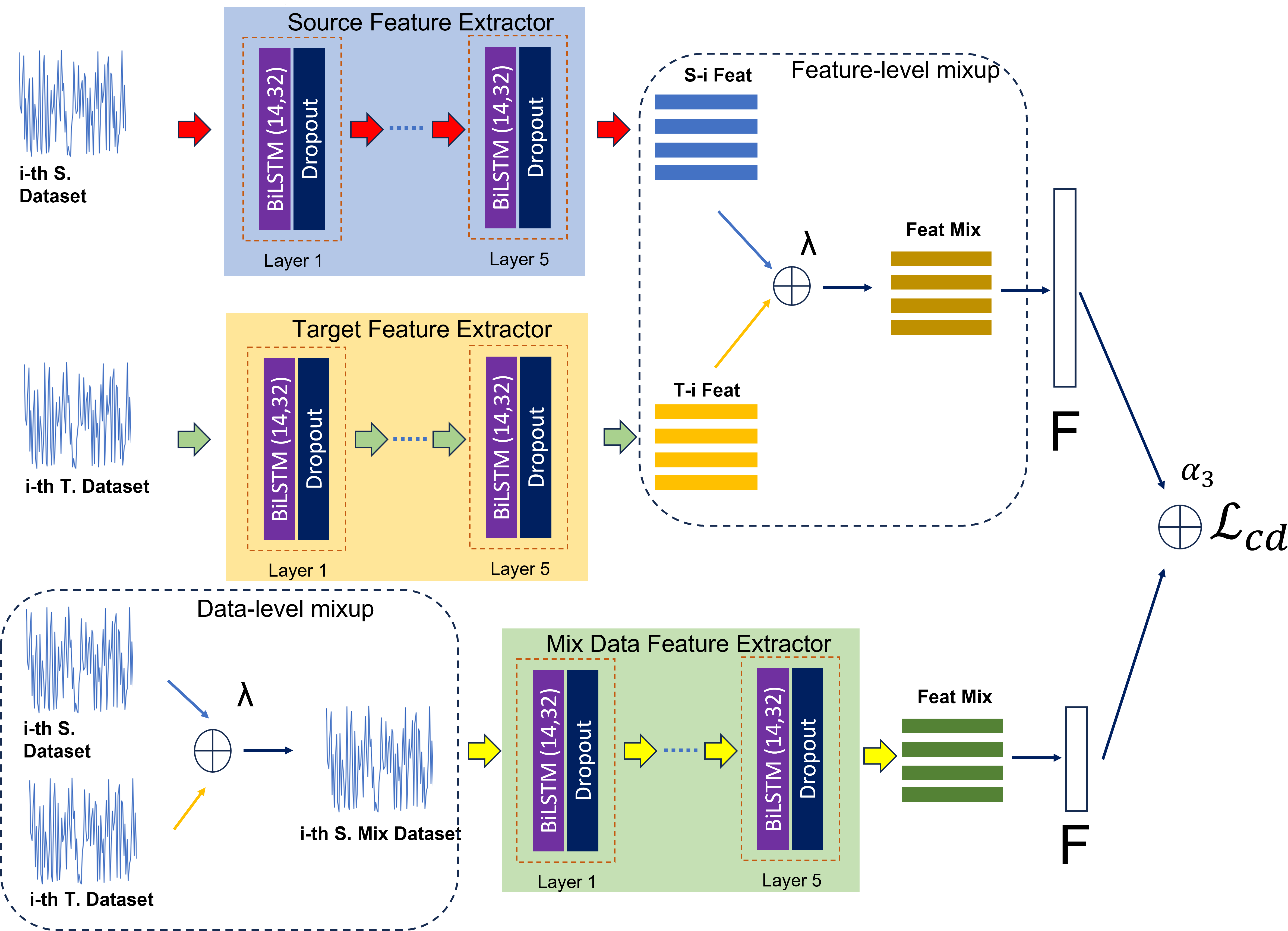}
    \caption{Intermediate Domain Training occurs with minimization of mix-up losses between the source domain samples and the target domain samples. The mix-up strategy dynamically interpolates samples of the input space and the feature space. The three backbone networks share the same network parameters.}
    \label{fig:intermediate domain}
\end{figure}

The mix-up samples and manifold mix-up samples are ordinarily learned in the supervised manner:
\begin{equation}
\mathcal{L}_{cd}=\mathbb{E}_{(x,y)\backsim\mathcal{\tilde{D}}_{S,T}}[l(f_{\phi}(g_{\psi}(\tilde{x})),\tilde{y})+l(f_{\phi}(\tilde{g}_{\psi}(x)),\tilde{y})]
\end{equation}
This learning procedure dampens the gap between the source and target domain by learning the auxiliary domain, namely the mix-up domain. The overall learning objective of the intermediate mix-up domain is formulated:
\begin{equation}
    \min_{\phi,\psi}\alpha_{3}\mathcal{L}_{cd}
\end{equation}
where $\alpha_{3}$ is a trade-off constant. An overview of the intermediate domain training phase is depicted in Fig. \ref{fig:intermediate domain}.

\subsection{Target Domain}
\begin{figure}
    \centering
    \includegraphics[width=1\linewidth]{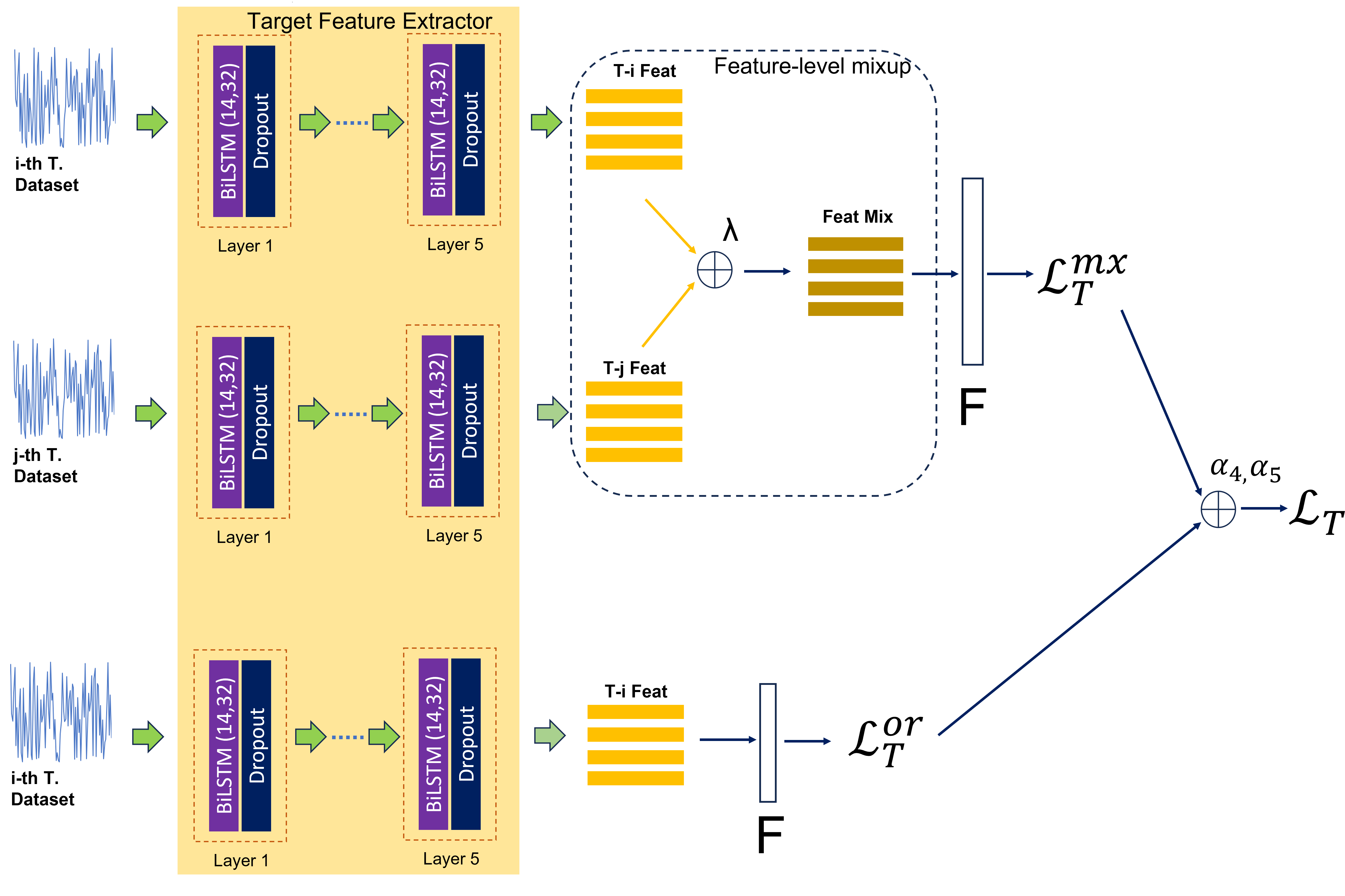}
    \caption{Target Domain Training is carried out by utilizing pseudo-labelled samples. Mixup samples are generated and optimized simultaneously with original pseudo-labelled samples.}
    \label{fig:target domain}
\end{figure}

The last step is to learn the target domain exploiting the pseudo-labels $\hat{y}_{T}$. That is, the self-learning mechanism takes place. 
\begin{equation}
\mathcal{L}_{T}^{or}=\mathbb{E}_{(x,y)\backsim\mathcal{D}_{T}}[l(f_{\phi}(g_{\psi}(x)),\hat{y})]
\end{equation}
In addition, the manifold mix-up strategy is implemented as done in the source domain for the consistency regularization. It generates mix-up samples $(f_{\phi}(\tilde{g}_{\psi}(x)),\tilde{y})$. The mix-up objective is defined:
\begin{equation}
  \mathcal{L}_{T}^{mx}=\mathbb{E}_{(x,y)\backsim\mathcal{D}_{T}}[l(f_{\phi}(\tilde{g}_{\psi}(x)),\tilde{y})]  
\end{equation}
Note that the consistency regularization step is achieved here because a model is forced to output the same output for both original samples and mixup samples. This strategy in turn reduces the detrimental effect of noisy pseudo labels. The overall learning objective for the target domain is formalized as follows:
\begin{equation}
\min_{\phi,\psi}\alpha_{4}\mathcal{L}_{T}^{or}+\alpha_{5}\mathcal{L}_{T}^{mx}
\end{equation}
where $\alpha_{4},\alpha_{5}$ are trade-off constants. Fig. \ref{fig:target domain} pictorially exhibits the target domain training process while the pseudo-code of MDAN is offered in Algorithm 1. 

\section{Experiments}
\subsection{Dataset}
Our numerical study is carried out with the C-MAPSS dataset \cite{Saxena2008DamagePM,Ragab2020AdversarialTL,Ragab2021ContrastiveAD} which presents the run-to-fail experiments of aircraft engines. It consists of four datasets, namely FD001-FD004 having different characteristics due to different working conditions, fault modes, life spans and number of engines. Table \ref{Table: dataset} summarizes their properties. This leads to 12 cases of unsupervised domain adaptation to be simulated in our study. We follow the same steps of data preparation as \cite{Ragab2020AdversarialTL,Ragab2021ContrastiveAD}. Sensor selection is done by choosing the ones depicting clear degradation trends from runs to failures since some sensors just provide constant trends being uninformative for RUL predictions. That is, common sensors across two domains are picked up for domain adaptations: S2, S3, S4, S7, S8, S9, S11, S12, S13, S14, S15, S17, S20 and S21. The min-max normalization steps are applied to ensure the same range across each sensors while the sliding window approach is applied to induce data samples from run-to-fail episodes with the window size of 30 and the step size of 1. The piece-wise linear RUL is adopted instead of the real RULs. That is, if the true RUL is larger than the maximal RUL, it is set to the maximal RUL. 

\begin{table}[]
\caption{THE C-MAPSS DATASET}
\centering
\begin{tabular}{c|cccc}
\hline
\textbf{Dataset}   & \textbf{FD001} & \textbf{FD002} & \textbf{FD003} & \textbf{FD004} \\ \hline
Working conditions & 1              & 6              & 1              & 6              \\ \hline
Fault modes        & 1              & 1              & 2              & 2              \\ \hline
Training engines   & 100            & 260            & 100            & 249            \\ \hline
Testing engines    & 100            & 259            & 100            & 248            \\ \hline
Training samples   & 17731          & 48558          & 21220          & 56815          \\ \hline
Testing samples    & 100            & 259            & 100            & 248            \\ \hline
\end{tabular}
\label{Table: dataset}
\end{table}

\subsection{Implementation Details}
MDAN's architecture is configured as with \cite{Ragab2021ContrastiveAD} to ensure fair comparisons. The feature extractor is implemented as a 5-layer biLSTM network where each layer comprises 32 nodes. The RUL predictor consists of three fully connected layers with 32-16-1 nodes respectively. The ReLU activation function is selected as the nonlinear activation function and the dropout regularization is applied to minimize the over-fitting problem. \textcolor{black}{The hyper-parameters of MDAN are detailed in Table \ref{table: hyper}}.

\begin{table}[]
\caption{\textcolor{black}{LIST OF HYPER-PARAMETERS}}
\centering
\scalebox{0.8}{%
\begin{tabular}{c|cccccccc}
\hline
\textbf{Dataset} & \textbf{Epoch} & \textbf{Batch size} & \textbf{Learning rate} & \textbf{$\alpha_{1}$} & \textbf{$\alpha_{2}$} & \textbf{$\alpha_{3}$} & \textbf{$\alpha_{4}$} & \textbf{$\alpha_{5}$} \\ \hline
FD001   & 100   & 256        & 3e-04      & 0.5     & 0.5     & 1       & 1       & 1       \\ \hline
FD002   & 75    & 256        & 3e-04      & 0.5     & 0.5     & 1       & 1       & 1       \\ \hline
FD003   & 150   & 256        & 3e-04      & 0.5     & 0.5     & 1       & 1       & 1       \\ \hline
FD004   & 175   & 256        & 3e-04      & 0.5     & 0.5     & 1       & 1       & 1       \\ \hline
MFD     & 15    & 512        & 1e-04      & 0.5     & 0.5     & 1       & 1       & 1       \\ \hline
\end{tabular}}
\label{table: hyper}
\end{table}

\subsection{Evaluation Metrics}
Two evaluation metrics, namely root mean square error (RMSE), and Score, are used to evaluate the performance of consolidated algorithms. The RMSE metric is formalized as follows:
\begin{equation}
    RMSE = \sqrt{\frac{1}{N}\sum_{i=1}^{N}(\hat{y}_i-y_i)}
\end{equation}
where $y_i,\hat{y}_i$ stand for the true RUL and the predicted RUL respectively. $N$ denotes the number of evaluation samples. Nevertheless, the RMSE metric does not consider the early RUL and late RUL predictions equally. As a matter of fact, the late RUL compromises the system significantly \cite{Ragab2021ContrastiveAD}. The score metric penalizes the late RUL predictions and is defined:
\textcolor{black}{\begin{align*}
   Score=\frac{1}{N}\sum_{i=1}^{N}Score_i; Score_i = \begin{cases}(\exp{(\frac{\hat{y}_i-y_i}{13}-1)}), & (\hat{y}_i<y_i)\\
    (\exp{(\frac{\hat{y}_i-y_i}{10}-1)}), & (\hat{y}_i>y_i)\end{cases}
\end{align*}}
Our algorithms is executed three times using different random seeds. The final results are obtained as the average of the three consecutive runs.
\subsection{Baseline Algorithms}
MDAN is compared with six prominent algorithms, contrastive adversarial domain adaptation (CADA) \cite{Ragab2021ContrastiveAD}, correlation alignment (CORAL) \cite{Sun2016CorrelationAF}, deep domain confusion (DDC) \cite{Tzeng2014DeepDC}, Wasserstein distance guided representation learning (WDGRL) \cite{Shen2017WassersteinDG}, adversarial discriminative domain adaptation (ADDA) \cite{Tzeng2017AdversarialDD}, deep domain adaptation (DDARUL) \cite{Costa2019RemainingUL}:
\begin{itemize}
    \item CADA is developed from a combination of adversarial domain adaptation \cite{Ganin2015DomainAdversarialTO} and constrastive learning framework via the InfoNCE loss. 
    \item CORAL directly addresses the gap between the source and target domains. 
    \item DDC implements the MMD distance loss to be minimized to align the source and target domains. 
    \item WDGRL relies on the Wasserstein distance for domain adaptation. 
    \item ADDA is developed from the GAN to discover similar target domain features. 
    \item DDARUL applies an LSTM feature extractor playing the minmax game with a domain classifier network. 
\end{itemize}
Numerical results of other algorithms are obtained from \cite{Ragab2021ContrastiveAD} because the numerical results under our computing platforms using their official source codes are inferior compared to published results.
\subsection{Numerical Results}
Table \ref{table: numerical results} reports numerical results of consolidated algorithms. MDAN demonstrates superior performances compared to other algorithms where it outperforms its counterparts with significant margins in 12 out of 12 domain adaptation cases. The finding is also consistent 
for both RMSE and score. Only in one case, FD002$\rightarrow$FD001, MDAN shows lower score than CADA but the difference is marginal. This finding confirms the advantage of MDAN in handling dynamic RUL predictions where labelled source information is transferred across to unlabelled target domain under time-series data samples. This is achieved by applying the mix-up strategy in creating the intermediate domain which effectively dampens the gap between the source and target domains. The last step is to apply the self-learning strategy in the target domain guided by the mix-up strategy. That is, it extracts the discriminative information of the target domain addressing the conditional distribution shifts. Another advantage of MDAN lies in its simplicity because it removes the presence of a domain discriminator when performing unsupervised domain adaptations. 

\begin{figure}
    \centering
    \includegraphics[width=1\linewidth]{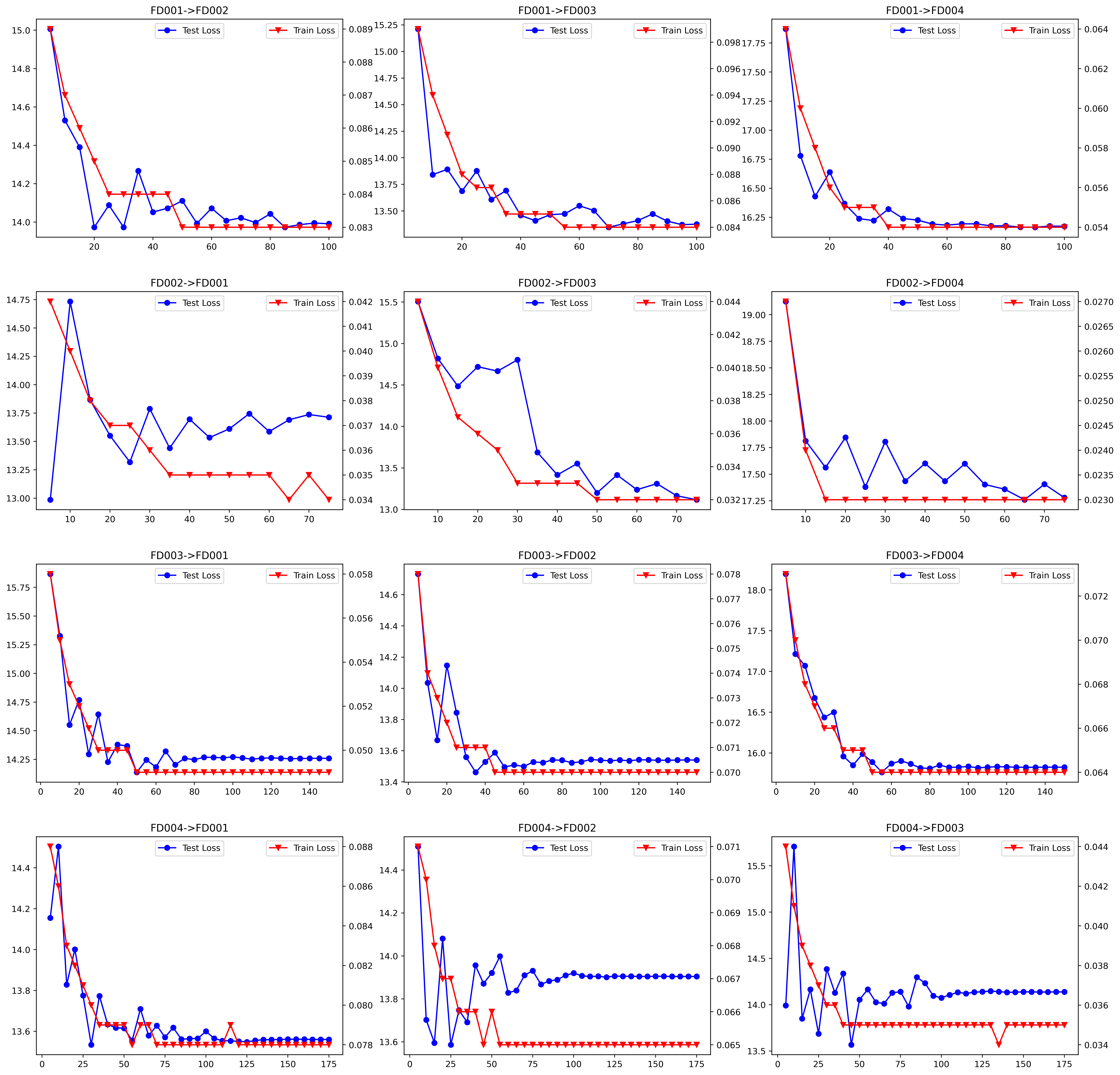}
    \caption{Train vs Test Loss}
    \label{fig:Train vs Test Loss}
\end{figure}

\begin{figure}
    \centering
    \includegraphics[width=1\linewidth]{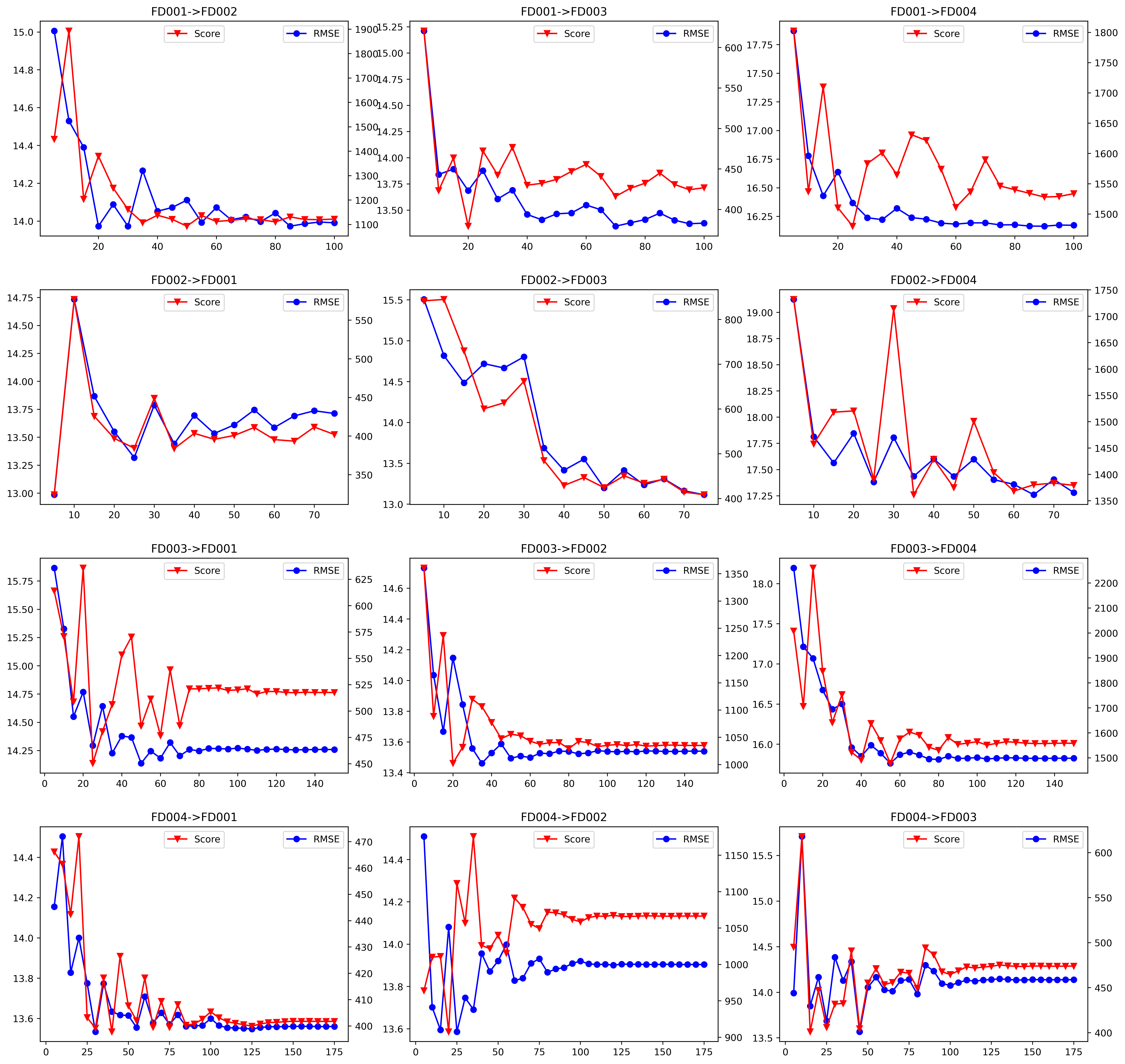}
    \caption{RMSE vs SCORE}
    \label{fig:RMSE vs SCORE}
\end{figure}

The training and validation losses of MDAN are portrayed in Fig. \ref{fig:Train vs Test Loss}. It is perceived that both losses converge and do not show any overfitting signs. Fig. \ref{fig:RMSE vs SCORE} depicts the evolution of RMSE and Score during the training process. Both RMSE and Score are stable and convergent during the training process.

\subsection{Analysis of Domain Discrepancy}
This subsection analyzes the advantage of MDAN in alleviating the discrepancies between the source domain and the target domain. This is done by enumerating the KL divergence loss $\mathcal{L}_{KL}(.)$ between the source domain and the target domain in the embedding space before and after the training process. Note that the KL divergence loss $\mathcal{L}_{KL}(.)$ is excluded from the training process. Numerical results are summarized in Table \ref{tab: KL divergence}. It is seen from Table \ref{tab: KL divergence} that the KL divergence losses decrease after the training process of MDAN in 11 of 12 domain adaptation cases. Only in one case, FD004$\rightarrow$FD003, the KL divergence loss slightly increases. This finding ensures the efficacy of MDAN in mitigating the discrepancies of the source domain and the target domain. It is worth mentioning that MDAN still beats other algorithms in both RMSE and Score for the FD004$\rightarrow$FD003.

\begin{table}[]
\caption{COMPARISON OF THE PROPOSED METHOD AGAINST STATE-OF-THE-ART APPROACHES}
\centering
\scalebox{0.5}{%
\begin{tabular}{l|ccccccc|ccccccc}
\hline
\multicolumn{1}{c}{Metric} & \multicolumn{7}{c}{RMSE}                                        & \multicolumn{7}{c}{SCORE}                                                  \\ \hline
\multicolumn{1}{c}{Method} & CORAL & WDGRL & DDC   & ADDA  & RULDDA & CADA  & MDAN           & CORAL  & WDGRL  & DDC     & ADDA   & RULDDA & CADA         & MDAN          \\ \hline
FD001 $\rightarrow$ FD002              & 22.85 & 21.46 & 44.05 & 31.26 & 24.08  & 19.52 & \textbf{13.99} & 2798   & 33160  & 5958    & 4865   & 2684   & 2122         & \textbf{1119} \\ \hline
FD001 $\rightarrow$ FD003              & 44.21 & 71.7  & 39.62 & 57.09 & 43.08  & 39.58 & \textbf{13.34} & 56991  & 15936  & 288061  & 32472  & 10259  & 8415         & \textbf{417}  \\ \hline
FD001 $\rightarrow$ FD004              & 50.03 & 57.24 & 44.35 & 56.66 & 45.7   & 31.23 & \textbf{16.12} & 52053  & 86139  & 156224  & 68859  & 26981  & 11577        & \textbf{1538} \\ \hline
FD002 $\rightarrow$ FD001              & 24.43 & 15.24 & 46.96 & 19.73 & 23.91  & 13.88 & \textbf{13.70} & 3590   & 157672 & 640     & 689    & 2430   & \textbf{351} & 393           \\ \hline
FD002 $\rightarrow$ FD003              & 42.66 & 41.45 & 39.87 & 37.22 & 47.26  & 33.53 & \textbf{13.35} & 23071  & 19053  & 62823   & 11029  & 12756  & 5213         & \textbf{426}  \\ \hline
FD002 $\rightarrow$ FD004              & 52.12 & 37.62 & 43.99 & 37.64 & 45.17  & 33.71 & \textbf{15.94} & 62852  & 52372  & 44872   & 16856  & 25738  & 15106        & \textbf{1319} \\ \hline
FD003 $\rightarrow$ FD001              & 40.33 & 36.05 & 39.95 & 40.41 & 27.15  & 19.54 & \textbf{14.18} & 4581   & 18307  & 25826   & 32451  & 2931   & 1451         & \textbf{507}  \\ \hline
FD003 $\rightarrow$ FD002              & 56.67 & 40.11 & 44.07 & 42.53 & 30.42  & 19.33 & \textbf{13.78} & 73026  & 32112  & 1012978 & 459911 & 6754   & 5257         & \textbf{1051} \\ \hline
FD003 $\rightarrow$ FD004              & 38.16 & 29.98 & 47.46 & 31.88 & 31.82  & 20.61 & \textbf{15.76} & 11407  & 296061 & 275665  & 82520  & 5775   & 3219         & \textbf{1552} \\ \hline
FD004 $\rightarrow$ FD001              & 51.44 & 42.01 & 41.55 & 37.81 & 32.37  & 20.10 & \textbf{14.12} & 154842 & 45394  & 162100  & 43794  & 13377  & 1840         & \textbf{422}  \\ \hline
FD004 $\rightarrow$ FD002              & 31.61 & 35.88 & 43.99 & 36.67 & 27.54  & 18.5  & \textbf{14.16} & 38095  & 38221  & 179243  & 23822  & 4937   & 4460         & \textbf{1096} \\ \hline
FD004 $\rightarrow$ FD003              & 30.44 & 18.18 & 44.47 & 23.59 & 23.31  & 14.49 & \textbf{14.26} & 6919   & 77977  & 1623    & 1117   & 1679   & 682          & \textbf{481}  \\ \hline
\end{tabular}}
\label{table: numerical results}
\end{table}

\begin{table}[]
\caption{KL-DIVERGENCE BETWEEN SOURCE AND TARGET DOMAIN BEFORE AND AFTER TRAINING}
\centering
\small
\begin{tabular}{c|c|c}
\hline
Scenarios     & Before Training & After Training \\ \hline
FD001 $\rightarrow$ FD002 & 0.096           & 0.082          \\ \hline
FD001 $\rightarrow$ FD003 & 0.091           & 0.056          \\ \hline
FD001 $\rightarrow$ FD004 & 0.104           & 0.070          \\ \hline
FD002 $\rightarrow$ FD001 & 0.095           & 0.083          \\ \hline
FD002 $\rightarrow$ FD003 & 0.086           & 0.052          \\ \hline
FD002 $\rightarrow$ FD004 & 0.076           & 0.045          \\ \hline
FD003 $\rightarrow$ FD001 & 0.080           & 0.056          \\ \hline
FD003 $\rightarrow$ FD002 & 0.054           & 0.043          \\ \hline
FD003 $\rightarrow$ FD004 & 0.068           & 0.065          \\ \hline
FD004 $\rightarrow$ FD001 & 0.091           & 0.066          \\ \hline
FD004 $\rightarrow$ FD002 & 0.077           & 0.077          \\ \hline
FD004 $\rightarrow$ FD003 & 0.071           & 0.097          \\ \hline
\end{tabular}
\label{tab: KL divergence}
\end{table}

\subsection{\textcolor{black}{Fault Diagnosis}}
This section evaluates the advantage of MDAN in performing the fault diagnosis of bearing machines using the MFD dataset \cite{Lessmeier2016ConditionMO,Ragab2021SelfsupervisedAD} where there exist three different classes, i.e., healthy, inner-bearing damage and outer-bearing damage. There are four different operating conditions referring to different rotational speeds, load torques and radial forces and leading to 12 domain adaptation cases. Unlike the RUL prediction cases, classification problems are handled here. Hence, we implement a threshold in our self-learning mechanism to prevent noisy pseudo labels. Our experiments follow the same setting as \cite{Ragab2021SelfsupervisedAD} where the window size of 5120 and the shifting size of 4096 are applied.

\begin{table}[]
\caption{\textcolor{black}{COMPARISON OF THE PROPOSED METHOD AGAINST SLARDA USING THE MFD DATASET}}
\centering
\small
\begin{tabular}{c|cc|cc|cc}
\hline
\multirow{2}{*}{Scenarios} & \multicolumn{2}{c|}{Source Only} & \multicolumn{2}{c|}{SLARDA} & \multicolumn{2}{c}{MDAN} \\ \cline{2-7} 
                           & Acc             & Stdv           & Acc               & Stdv    & Acc              & Stdv  \\ \hline
a $\rightarrow$ b                      & 30.18           & 0.00           & 30.96             & 0.08    & \textbf{79.81}   & 0.02  \\ \hline
a $\rightarrow$ c                      & 30.18           & 0.00           & 30.54             & 0.00    & \textbf{89.57}   & 0.00  \\ \hline
a $\rightarrow$ d                      & 45.56           & 0.00           & \textbf{84.78}    & 0.29    & 28.14            & 0.08  \\ \hline
b $\rightarrow$ a                      & 35.53           & 11.97          & 27.85             & 0.16    & \textbf{70.64}   & 0.08  \\ \hline
b $\rightarrow$ c                      & 30.18           & 0.00           & 29.13             & 0.09    & \textbf{71.19}   & 0.47  \\ \hline
b $\rightarrow$ d                      & 45.56           & 0.00           & \textbf{99.89}    & 0.08    & 17.96            & 0.00  \\ \hline
c $\rightarrow$ a                      & 35.53           & 11.95          & 37.93             & 0.60    & \textbf{89.15}   & 0.00  \\ \hline
c $\rightarrow$ b                      & 30.18           & 0.00           & 28.86             & 0.28    & \textbf{87.54}   & 0.07  \\ \hline
c $\rightarrow$ d                      & 45.56           & 0.00           & \textbf{89.82}    & 0.31    & 34.88            & 0.42  \\ \hline
d $\rightarrow$ a                      & 30.17           & 0.03           & 27.16             & 0.09    & \textbf{30.50}   & 0.00  \\ \hline
d $\rightarrow$ b                      & 30.18           & 0.00           & 23.94             & 0.14    & \textbf{30.50}   & 0.00  \\ \hline
d $\rightarrow$ c                      & 30.18           & 0.00           & \textbf{30.94}    & 0.10    & 30.50            & 0.00  \\ \hline
\end{tabular}
\label{table: mfd results}
\end{table}

\textcolor{black}{MDAN is compared with Source only without any domain adaptation, and SLARDA \cite{Ragab2021SelfsupervisedAD} where comparisons are made using their publicly available codes}. The same network architecture, namely the 1-D CNN with 5 layers, is used to assure fair comparisons. The learning rate of $3e-4$ is implemented with the dropout of $0.001$ and the pseudo-label threshold of $0.9$ where they are found using the grid search approach. Numerical results are reported in Table \ref{table: mfd results} and taken from the mean of 5 consecutive runs with different random seeds. 

It is perceived from Table \ref{table: mfd results} that MDAN outperforms SLARDA with noticeable margins in 8 of 12 domain adaptation cases of the MFD dataset. This finding confirms that the mix-up-based domain adaptation strategy produces higher classification rates than the adversarial-based domain adaptation strategy as SLARDA. MDAN is inferior to SLARDA in 4 cases, a$\rightarrow$d, b$\rightarrow$d, c$\rightarrow$d, d$\rightarrow$c. Note that the domain alignment procedure of MDAN is done with the absence of the domain discriminator, thus reducing memory burdens.

\begin{figure}
    \centering
    \includegraphics[width=1\linewidth]{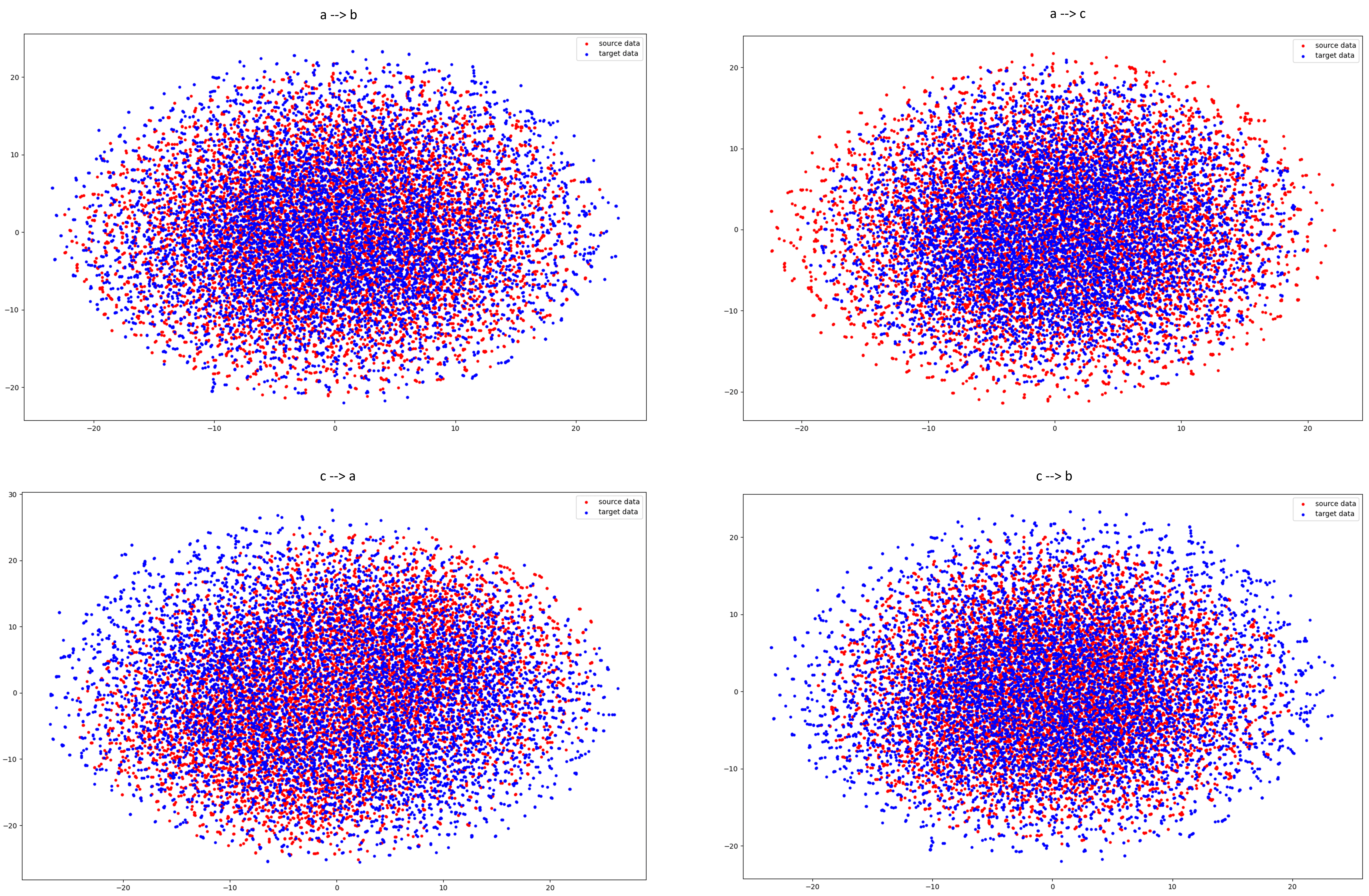}
    \caption{\textcolor{black}{T-SNE Plot of MFD Dataset Before Training}}
    \label{fig:TSNE MFD BFR}
\end{figure}

\begin{figure}
    \centering
    \includegraphics[width=1\linewidth]{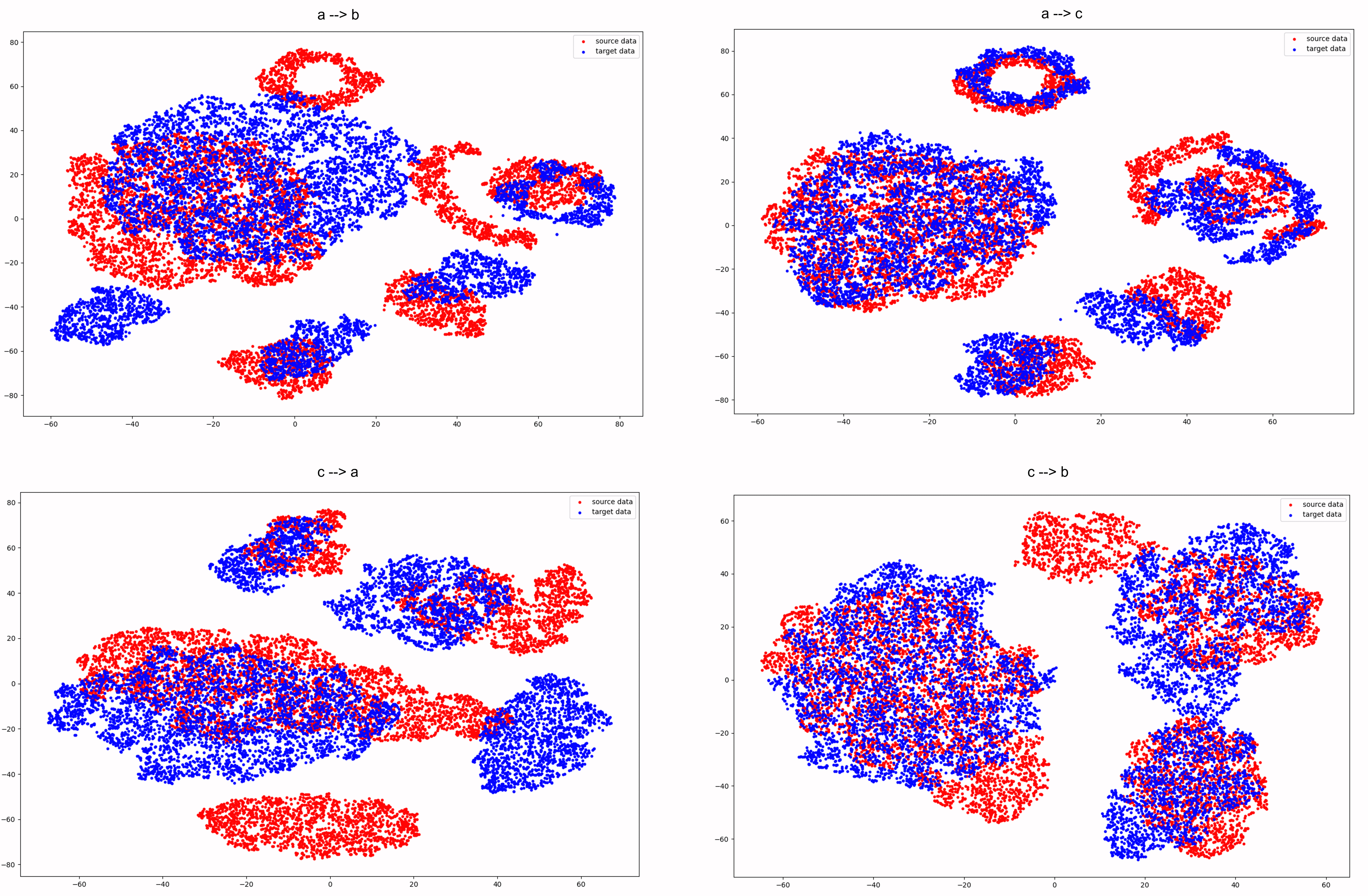}
    \caption{T-SNE Plot of MFD Dataset After Training}
    \label{fig:TSNE MFD}
\end{figure}

\section{Ablation Study}
This section describes our ablation study across all domain adaptation cases. Table \ref{Table: ablation study} reports our numerical results. 
\subsection{Source Only w/o Self-Training Mechanism}
Our first ablation study is to check the performance of MDAN when trained in the source domain only. That is, the intermediate domain and the target domain are absent. It is perceived from Table \ref{Table: ablation study} that MDAN's performances are significantly compromised under this setting. The underlying rationale is found in the fact that the domain shift is not resolved in this scenario. In other words, no domain adaptation strategies take place in this setting.
\subsection{Source + Intermediate w/o Target Domain Training}
Next, we evaluate the performance of MDAN under the source and intermediate domain training only without any target domain training. It is discovered that the target domain training plays vital role to MDAN's performances. That is, MDAN fails to extract the discriminative information of the target domain with the absence of the target domain training. The self-learning mechanism with pseudo-labels contributes toward the performance improvements while the mix-up strategy plays a regularization role to reject noisy pseudo labels.
\subsection{The absence of Mixup}
MDAN is evaluated with the absence of mix-up strategies. This setting turns out to be disadvantageous to MDAN as shown by significant performance drops. This is caused by the absence of domain alignment where the intermediate domain is created to reduce the gaps of the source domain and the target domain. The absence of mixup also implies the removals of consistency regularization in the source domain and the target domain leading to performance degradation. Our ablation study substantiates the importance of MDAN's learning modules.  
\subsection{w/o Self-Supervised Learning}
This subsection outlines the advantage of self-supervised learning via the controlled reconstruction learning strategy for MDAN. It is evident that the self-supervised learning strategy boosts the performance of MDAN, i.e., the performance of MDAN compromises in 11 of 12 cases. Only in one case, FD004$\rightarrow$FD001, returns the opposite case yet the difference is trivial. The self-supervised learning strategy is capable of extracting general features which can be easily transferred to another domain, thus overcoming the domain shift problem.

\begin{table*}[]
\caption{ABLATION STUDY OF THE PROPOSED APPROACH}
\centering
\scalebox{0.5}{%
\begin{tabular}{lcccccccccc}
\hline
\multicolumn{1}{c}{Metric}                  & \multicolumn{5}{c}{RMSE}                                                                                                    & \multicolumn{5}{c}{SCORE}                                                           \\ \hline
\multicolumn{1}{c}{\multirow{2}{*}{Method}} & \multirow{2}{*}{Source Only} & Source +     & Source + Target & w/o Self-      & \multicolumn{1}{c|}{\multirow{2}{*}{MDAN}} & Source Only & Source +     & Source + Target & w/o Self-    & \multirow{2}{*}{MDAN} \\
\multicolumn{1}{c}{}                        &                              & Intermediate & (w/o mixup)     & supervised     & \multicolumn{1}{c|}{}                      &             & Intermediate & (w/o mixup)     & supervised   &                       \\ \hline
\multicolumn{1}{l|}{FD001 $\rightarrow$ FD002}          & 20.77                        & 23.58        & 20.12           & 15.56          & \multicolumn{1}{c|}{\textbf{13.99}}        & 8077        & 2652         & 5778            & 1247         & \textbf{1119}         \\ \hline
\multicolumn{1}{l|}{FD001 $\rightarrow$ FD003}          & 50.61                        & 79.35        & 45.76           & 14.30          & \multicolumn{1}{c|}{\textbf{13.34}}        & 33093       & 226406       & 18927           & 681          & \textbf{417}          \\ \hline
\multicolumn{1}{l|}{FD001 $\rightarrow$ FD004}          & 33.81                        & 69.60        & 31.68           & 21.40          & \multicolumn{1}{c|}{\textbf{16.12}}        & 11424       & 344526       & 9136            & 2578         & \textbf{1538}         \\ \hline
\multicolumn{1}{l|}{FD002 $\rightarrow$ FD001}          & 15.77                        & 15.10        & 16.34           & 24.27          & \multicolumn{1}{c|}{\textbf{13.70}}        & 743         & 898          & 1001            & 1770         & \textbf{393}          \\ \hline
\multicolumn{1}{l|}{FD002 $\rightarrow$ FD003}          & 39.97                        & 59.45        & 40.65           & 28.67          & \multicolumn{1}{c|}{\textbf{13.35}}        & 11552       & 34782        & 12654           & 3923         & \textbf{426}          \\ \hline
\multicolumn{1}{l|}{FD002 $\rightarrow$ FD004}          & 38.84                        & 59.26        & 38.42           & 40.64          & \multicolumn{1}{c|}{\textbf{15.94}}        & 18032       & 87352        & 21294           & 62450        & \textbf{1319}         \\ \hline
\multicolumn{1}{l|}{FD003 $\rightarrow$ FD001}          & 30.50                        & 45.94        & 28.94           & 19.14          & \multicolumn{1}{c|}{\textbf{14.18}}        & 7119        & 8252         & 5547            & 668          & \textbf{507}          \\ \hline
\multicolumn{1}{l|}{FD003 $\rightarrow$ FD002}          & 40.40                        & 54.98        & 29.10           & 17.47          & \multicolumn{1}{c|}{\textbf{13.78}}        & 234785      & 64655        & 61745           & 1472         & \textbf{1051}         \\ \hline
\multicolumn{1}{l|}{FD003 $\rightarrow$ FD004}          & 24.24                        & 46.74        & 23.13           & 19.01          & \multicolumn{1}{c|}{\textbf{15.76}}        & 18927       & 26299        & 29673           & 1813         & \textbf{1552}         \\ \hline
\multicolumn{1}{l|}{FD004 $\rightarrow$ FD001}          & 32.90                        & 20.60        & 39.35           & \textbf{13.62} & \multicolumn{1}{c|}{14.12}                 & 19225       & 855          & 122741          & \textbf{305} & 422                   \\ \hline
\multicolumn{1}{l|}{FD004 $\rightarrow$ FD002}          & 26.52                        & 32.53        & 30.98           & 44.53          & \multicolumn{1}{c|}{\textbf{14.16}}        & 14406       & 7135         & 78435           & 73940        & \textbf{1096}         \\ \hline
\multicolumn{1}{l|}{FD004 $\rightarrow$ FD003}          & 19.61                        & 19.04        & 21.92           & 23.18          & \multicolumn{1}{c|}{\textbf{14.26}}        & 2941        & 527          & 2508            & 1448         & \textbf{481}          \\ \hline
\end{tabular}}
\label{Table: ablation study}
\end{table*}

\section{T-SNE Analysis}
\textcolor{black}{This section discusses the embedding qualities of MDAN to evaluate whether it is capable of aligning the source domain and the target domain. This is done using the T-SNE plots where Fig. \ref{fig:TSNE MFD} displays the t-sne plots of 4 domain adaptation cases of the MFD dataset, a$\rightarrow$b, a$\rightarrow$c, c$\rightarrow$a, c$\rightarrow$b after performing the training process while Fig. \ref{fig:TSNE MFD BFR} exhibits the t-sne plots of the 4 domain adaptation cases of the MFD dataset before the training process begins. Initially, there do not exist any cluster or class structures at all in the T-SNE plots indicating model confusions as shown in Fig. \ref{fig:TSNE MFD BFR}. After completing the training process, it is seen from Fig. \ref{fig:TSNE MFD} that MDAN attains decent embedding where the source-domain samples and the target-domain samples can be successfully aligned. That is, samples of the same class are mapped in the similar regions regardless of their domains. This finding substantiates our claims that MDAN can dampen the gaps between the source domain and target domain.} 

\section{Limitations}
Although MDAN outperforms prior arts in almost all cases, the following explains several drawbacks of MDAN:
\begin{itemize}
    \item MDAN still ignores the issue of data privacy when performing domain adaptations. That is, it calls for source domain samples which might be unavailable due to the privacy reasons. 
    \item MDAN still excludes the issue of lifelong learning which can be valid for the cases of RUL predictions due to continous operating conditions. 
    \item MDAN still assumes the closed-set domain adaptation problems where the source and target labels are exactly the same. In practise, this condition may not hold. 
    \item MDAN also does not explore the possibility of multi-source domains which can be used to boost numerical results.
\end{itemize}

\section{Conclusions}
Mixup Domain Adaptation (MDAN) is proposed in this paper for dynamic RUL predictions where a model is suppossed to generalize well in the unlabelled target domain given the labelled source domain. MDAN characterizes a three-staged learning process. At first, the labelled source domain is learned in the supervised manner guided with the mixup regularization strategy and the self-supervised learning strategy to prevent the supervision collapse problem. The second step establishes the intermediate domain exploiting the mixup samples of the source domain samples and the target domain samples to reduce the gap between the two domain. The last step is done with the self-learning strategy coupled with the mixup regularization strategy to extract the discriminative information of the target domain and to prevent the noisy pseudo labels. Our experiments demonstrates the efficacy of MDAN where it beats prominent domain adaptation algorithms with significant margins in 12 out of 12 cases. We also supplement this finding with other experiments using the bearing machine dataset, the MFD dataset. It is shown that MDAN outperforms prior art in 8 of 12 domain adaptation cases. In addition, our ablation study confirms the contributions of each learning component of MDAN. 

In practise, the issue of privacy plays vital role in many applications. This implies the absence of source domain samples for domain adaptations. Our future work is devoted to study the problem of source-free time-series domain adaptations where only a pretrained source model is required without any access to the source domain samples.

\section{\textcolor{black}{List of Acronyms}}
\begin{acronym}[DDARUL] 
\acro{ADDA}{Adversarial Discriminative Domain Adaptation}
\acro{AI}{Artificial Intelligence}
\acro{biLSTM}{bi-directional Long Short-Term Memory}
\acro{BN}{Batch Normalization}
\acro{CADA}{Contrastive Adversarial Domain Adaptation}
\acro{CNN}{Convolutional Neural Network}
\acro{CORAL}{Correlation Alignment}
\acro{DBN}{Deep Belief Network}
\acro{DDARUL}{Deep Domain Adaptation}
\acro{DDC}{Deep Domain Confusion}
\acro{GAN}{Generative Adversarial Network}
\acro{I.I.D}{Independent and Identically Distributed}
\acro{LSTM}{Long Short-Term Memory}
\acro{MDAN}{Mixup Domain Adaptation}
\acro{MMD}{Maximum Mean Discrepancy}
\acro{RMSE}{Root Mean Square Error}
\acro{RUL}{Remaining Useful Life}
\acro{SLARDA}{Self-supervised AutoRegressive Domain Adaptation}
\acro{T-SNE}{t-distributed Stochastic Neighbor Embedding}
\acro{UDA}{Unsupervised Domain Adaptation}
\acro{WDGRL}{Wasserstein Distance Guided Representation Learning}
\end{acronym}


 \bibliographystyle{elsarticle-num} 
 \bibliography{cas-refs}





\end{document}